\documentclass[conference, review]{IEEEtran}
\IEEEoverridecommandlockouts
\usepackage{cite}
\usepackage{amsmath,amssymb,amsfonts}
\usepackage{algorithmic}
\usepackage{graphicx}
\usepackage{textcomp}
\usepackage{xcolor}
\usepackage{hyperref}
\usepackage{comment}
\def\BibTeX{{\rm B\kern-.05em{\sc i\kern-.025em b}\kern-.08em
    T\kern-.1667em\lower.7ex\hbox{E}\kern-.125emX}}
\begin{document}

\newcommand{\gina}[1]{\textit{\textcolor{purple}{ Gina: #1}}}
\newcommand{\todo}[1]{\textit{\textcolor{red}{#1}}}
\newcommand{\gapi}[1]{\textit{\textcolor{green}{ This text is copied from the other paper: (#1)}}}
\newcommand{\tome}[1]{\textit{\textcolor{blue}{ Tome: #1}}}

\title{TransOpt: Transformer-based Representation Learning for Optimization Problem Classification\\}

\author{
\IEEEauthorblockN{1\textsuperscript{st} Gjorgjina Cenikj}
\IEEEauthorblockA{
\textit{Computer Systems Department} \\
\textit{Jo\v{z}ef Stefan Institute}\\
\textit{Jo\v{z}ef Stefan International}\\ 
\textit{Postgraduate School}\\
Ljubljana, Slovenia \\
gjorgjina.cenikj@ijs.si}
\and
\IEEEauthorblockN{2\textsuperscript{nd} Gašper Petelin}
\IEEEauthorblockA{\textit{Computer Systems Department} \\
\textit{Jo\v{z}ef Stefan Institute}\\
\textit{Jo\v{z}ef Stefan International}\\ 
\textit{Postgraduate School}\\
Ljubljana, Slovenia \\
gasper.petelin@ijs.si}
\and
\IEEEauthorblockN{3\textsuperscript{rd} Tome Eftimov}
\IEEEauthorblockA{\textit{Computer Systems Department} \\
\textit{Jo\v{z}ef Stefan Institute}\\
Ljubljana, Slovenia \\
tome.eftimov@ijs.si}
}

\maketitle

\begin{abstract} We propose a representation of optimization problem instances using a transformer-based neural network architecture trained for the task of problem classification of the 24 problem classes from the Black-box Optimization Benchmarking (BBOB) benchmark. We show that transformer-based methods can be trained to recognize problem classes with accuracies in the range of 70\%-80\% for different problem dimensions, suggesting the possible application of transformer architectures in acquiring representations for black-box optimization problems.

\end{abstract}

\begin{IEEEkeywords}
single-objective continuous optimization, representation learning, problem landscape features
\end{IEEEkeywords}

\section{Introduction}
The representation of optimization problems in terms of numerical features, often referred to as problem landscape features, is essential for the Machine Learning based analyses of the similarity and representativeness of problem instances~\cite{urban_ela_not_invariant,selector,kate_space_filling_instances}, automated algorithm selection and configuration. 
Several types of features capturing properties of single-objective continuous optimization problems have been proposed, which can be broadly categorized as explicitly defined features (such as those based on Fitness Landscape Analysis (FLA)~\cite{fla}, Exploratory Landscape Analysis (ELA)~\cite{ela} and Topological Landscape Analysis (TLA)~\cite{tla}), and feature-free approaches~\cite{dl_feature_free, doe2vec}. 
However, FLA features require human effort to be computed, while the ELA features can be computationally expensive to compute for high dimensional problems, have been shown to be sensitive to the sample size and sampling method~\cite{ela_sensitive_to_sampling}, and are not invariant to transformations such as scaling and shifting of the optimization problem~\cite{urban_ela_not_invariant}. 
%Feature-free approaches using recent advances in deep learning have also been explored for the prediction of FLA features from samples of the optimization problem~\cite{dl_feature_free}.

Motivated by the success of transformer-based models~\cite{attention} in various fields and applications, in this paper, we explore the possibility of generating vectorized representations of optimization problems through the use of transformer models applied to samples of the optimization problem. In particular, we train a transformer-based model architecture for classifying optimization problem instances from the Black-box Optimization Benchmarking (BBOB)~\cite{bbob,coco} suite into one of the 24 problem classes.

\section{Methodology}
\label{methodology}

The Black-box Optimization Benchmarking (BBOB)~\cite{bbob,coco} suite contains 24 single-objective optimization problem classes. Each problem class can further have multiple problem instances, which are a transformation of the original problem class. We use the first 999 instances of each problem class to train the transformer model to predict one of the 24 problem classes. We use problem classes of dimensions 3 and 20.

The first step of the proposed methodology involves the generation of samples from which the problem representations are calculated. The samples are obtained using Latin Hypercube Sampling, and we explore sample sizes of 50$d$ and 100$d$, where $d$ is the problem dimensionality. 
The objective values of the samples (y-values) are scaled to be in the range [0,1], while the candidate solutions (x-values) are kept in their original range of [-5,5], since this range is fixed in the BBOB.

\begin{figure}
    \centering
    \includegraphics[width=0.40\linewidth]{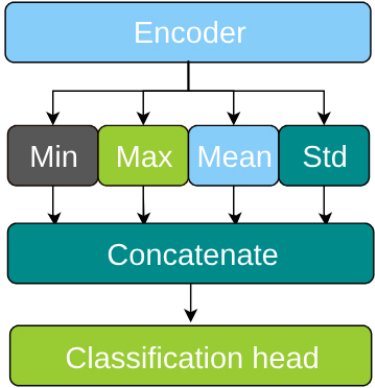}
    \caption{Model architecture}
    \label{fig:architecture}
\end{figure}

Figure~\ref{fig:architecture} shows the model architecture. The input to the model are the samples from the optimization problems. 
A single training instance of the ML model is a matrix of shape $[ s, d + 1 ]$, where $s$ is the number of problem instance samples, and $d$ is the dimensionality of the problem. In this case, the second dimension of the input matrix is $d + 1$, since the value of the objective function is also included.
The encoder part of the transformer model produces an embedding of size $e$, which is a specified model parameter, for each of the given samples from the optimization problem, i.e. it outputs a matrix of the shape $[s, e]$. In order to obtain a single, flat representation of the problem, we calculate several descriptive statistics on the matrix obtained from the encoder. In particular, we calculate the minimum, maximum, mean, and standard deviation of the representations of the samples produced by the encoder. Concatenating the vectors obtained using each descriptive statistic, we obtain the representation of the problem, which is then fed into a classification head, which produces the class of the problem.
The classification head contains a linear layer with a Rectified Linear Unit activation, a dropout layer, and a linear layer which performs the classification into 24 problem classes.

The model is trained with a cross-entropy loss, the Adam optimizer and a maximal learning rate of 0.001. The training is executed for a maximum of 200 epochs, with an early stopping mechanism that prevents overfitting by terminating the training process if the validation loss does not observe a decrease of at least 0.001 for five epochs. We use stratified 10-fold cross-validation to evaluate the model.

\section{Results}
\label{results}
\begin{figure}
    \centering
    \includegraphics[width=\linewidth]{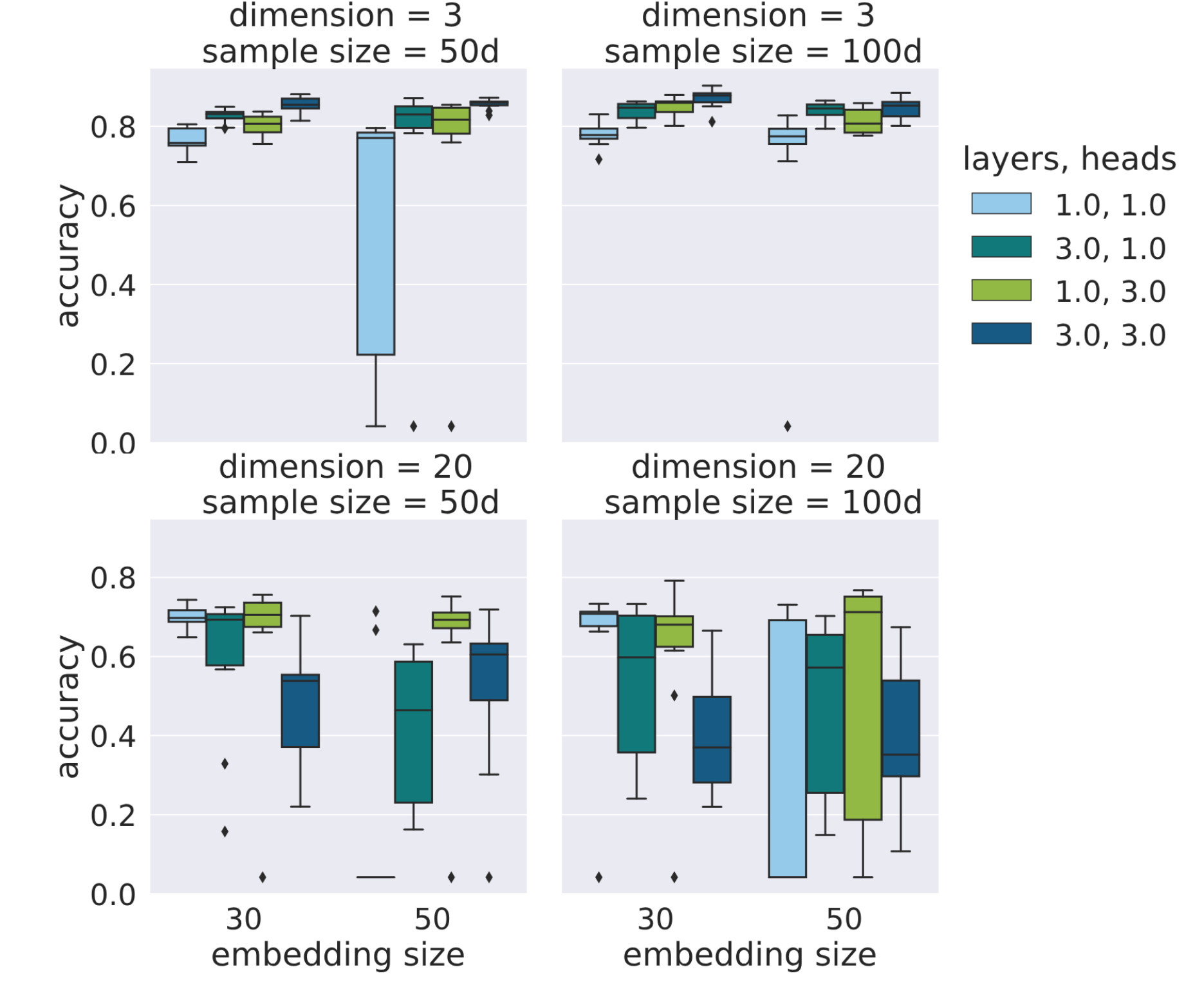}
    \caption{Accuracy obtained with different model parameters}
    \label{fig:conf_accuracy}
\end{figure}

In order to find a set of reasonable model parameters for further analysis of the architecture, we first ran an initial analysis of the impact of different model parameters on the obtained problem classification accuracy. Figure~\ref{fig:conf_accuracy} shows the accuracy obtained in the problem classification task with different parameters for problems of dimension 3 and 20.
In particular, we tune the number of heads and layers in the transformer architecture, as well as the dimension of the generated embeddings. 
The results indicate that increasing the number of heads and layers has a positive impact on the classification accuracy for lower dimensions, however, the same does not always hold for higher dimensions.
Increasing the sample size does not seem to substantially impact the classification accuracy for problems of lower dimensions, however, for larger dimensions, the model training is not successful with a sample size of 100$d$.
In general, smaller models with lower values of the investigated parameters seem to provide satisfactory results for all dimensions. The best performing configuration is (sample size = 50$d$, embedding size = 30, heads = 1, layers = 1), providing accuracies in the range 70\%-80\% for problems of different dimensions.

\section{Conclusion}
\label{conclusion}
We propose a novel representation for single objective optimization problem instances, which uses a transformer model trained for the task of problem classification of samples of the problem instances. The proposed model achieves accuracies around 70\%-80\% for different problem dimensions, indicating its potential for representing high-dimensional problems and possible utilization as end-to-end algorithm selection models.

\section{Acknowledgements}
Funding in direct support of this work: Slovenian Research Agency: research  core  funding  No. P2-0098, young researcher grants No. PR-12393 to GC and No. PR-11263 to GP, and project No. N2-0239 to TE and project No. J2-4460.
%Funding in direct support of this work: Slovenian Research Agency: research  core  funding  No. P2-0098, young researcher grants No. PR-12393 to GC and No. PR-11263 to GP, projects No. N2-0239 to TE and No. J2-4460 to PK, and a bilateral project between Slovenia and France grant No. BI-FR/23-24-PROTEUS-001 (PR-12040). Our work is also supported by ANR-22-ERCS-0003-01 project VARIATION.
\bibliographystyle{IEEEtran}
\bibliography{references}

\end{document}